\documentclass[letterpaper, 10 pt, conference]{packages/ieeeconf}

\IEEEoverridecommandlockouts                              

\usepackage{times}
\usepackage{epsfig}
\usepackage{graphicx}
\usepackage{amsmath}
\usepackage{amssymb}
\usepackage{subfigure} 
\usepackage[ruled]{algorithm}
\usepackage{algpseudocode}
\usepackage{ctable}
\usepackage{url}
\usepackage[bookmarks=true]{hyperref}
\usepackage[export]{adjustbox}
\usepackage{import}
\usepackage{pgf}
\usepackage{setspace}

\makeatletter
\let\NAT@parse\undefined
\makeatother
\usepackage[numbers]{natbib}

\usepackage{packages/sparo_acronyms}
\usepackage{packages/sparo_math}
\usepackage{packages/sparo_SIunits}
\usepackage{packages/sparo_misc}
\usepackage{multirow}

\usepackage{soul,color}
\usepackage{verbatim} 

\usepackage{xcolor}
\newcommand{\bl}[1]{{\textcolor{blue}{#1}}}

\definecolor{gl}{HTML}{008000}

\DeclareMathOperator*{\argmin}{argmin}

\usepackage[symbol]{footmisc}

\title{\LARGE \bf
    ReFeree: Radar-based efficient global descriptor \\ using a Feature and Free space for Place Recognition

}

\author{
    Byunghee Choi$^{1*}$, Hogyun Kim$^{1*}$, and Younggun Cho$^{1\dagger}$
    \thanks{
        $^{1*}$Byunghee Choi, $^{1*}$Hogyun Kim and $^{1\dagger}$Younggun Cho are with the Electrical and Computer Engineering, Inha University, Incheon, South Korea {\tt\small [bhbhchoi, hg.kim]@inha.edu, yg.cho@inha.ac.kr}
        , (*) represents equal contribution. 
        \newline
        This work was supported by the National Research Foundation of Korea (NRF) grant funded by the Korea government (MSIP) (No.2022R1A4A3029480 and RS-2023-00302589) and by the Institute of Information $\&$ communications Technology Planning $\&$ Evaluation (IITP) grant funded by the Korea government (MSIT) (No.2022-0-00448).
    }%
}

\begin{document}

\maketitle
\thispagestyle{empty}
\pagestyle{empty}

\begin{abstract} 
Radar is highlighted for robust sensing capabilities in adverse weather conditions (e.g. dense fog, heavy rain, or snowfall). 
In addition, Radar can cover wide areas and penetrate small particles.
Despite these advantages, Radar-based place recognition remains in the early stages compared to other sensors due to its unique characteristics such as low resolution, and significant noise.
In this paper, we propose a Radar-based place recognition utilizing a descriptor called \textit{ReFeree} using a feature and free space.
Unlike traditional methods, we overwhelmingly summarize the Radar image (i.e. 361.5KB $\rightarrow$ 528B).
Despite being lightweight, it contains semi-metric information and is also outstanding from the perspective of place recognition performance.
For concrete validation, we test a single session from the MulRan dataset and a multi-session from the Oxford Offroad Radar, Oxford Radar RobotCar, and the Boreas dataset.
The supplementary materials of our place recognition and SLAM are available at \url{https://sites.google.com/view/radar-referee}.
\end{abstract} 
\section{Introduction}
Place recognition is essential to autonomous vehicle missions (e.g. \ac{SLAM} or safety navigation).
Accurate place recognition enables correcting accumulated drift errors in a consistent map.
However, as the weather deteriorates, the decline in place recognition performance of most sensors becomes inevitable.

Instead of diverse sensors such as a camera \cite{arandjelovic2016netvlad}, \ac{LiDAR} \cite{kim2018scan}, or \ac{Sonar} \cite{kim2023robust} we propose a \ac{Radar}-based place recognition called \textit{ReFeree} using a feature and free space.
\ac{Radar} can overcome the harsh weather and boast a robust perception performance.
In addition, \ac{Radar} can penetrate small particles, and sensor data is impervious to occlusion.
Recently, \ac{Radar}-based place recognition studies have shown remarkable results \cite{kim2020mulran, jang2023raplace, gadd2024open}, but reliable localization remains challenging in various situations.
Also, the low resolution and significant noise in sensor data need to be solved.

\begin{figure}[t]
	\centering
	\def\width{0.48\textwidth}%
    {%
		\includegraphics[clip, trim= 70 240 70 0, width=\width]{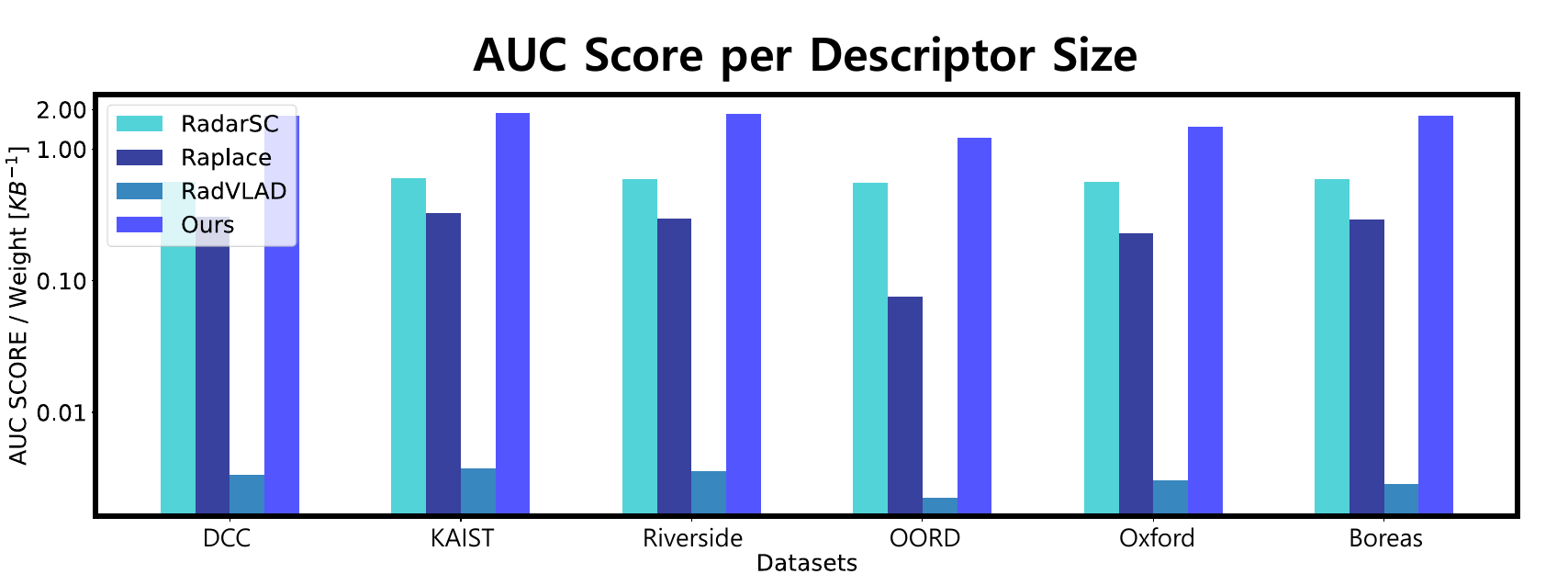}
	}
        \vspace{-0.5cm}
        \caption{Top is
                 The bottom is the Area Under the ROC Curve (AUC) score per descriptor's weight in single-session (S) \cite{kim2020mulran}  and multi-session datasets (M) \cite{kim2020mulran, barnes2020oxford, burnett2023boreas}.
                 We compare the proposed method with \textit{Radar Scan Context} (RadarSC) \cite{kim2020mulran}, \textit{Raplace} \cite{jang2023raplace} and \textit{Open-RadVLAD} (RadVLAD) \cite{gadd2024open} used in \ac{Radar}.
                 }
        \vspace{-0.5cm}
	
 \label{fig:main}
\end{figure}

The main contributions of this work are the following:
\begin{itemize}
    \item \textbf{Effective and Lightweight Description:} We propose an efficient three-step description.
                                                          First, we utilize prominent features from \ac{Radar} image to minimize noise.
                                                          Second, we check for areas where features have not been extracted, namely free space (i.e. empty space).
                                                          Finally, we incorporate free space along the range axis to generate a lightweight 1D vector.
                                                          Therefore, the elements forming each vector encapsulate the number of pixels, excluding strong signals.
                                                          Also, since we use the \ac{Radar} image without transforming it into different properties, the proposed method includes the semi-metric information.
                                                          Our descriptor is overwhelmingly lightweight compared to other descriptors.
                                                          As shown in \figref{fig:main}, our descriptor demonstrates superior performance relative to its size.
                                         
    \item \textbf{Various Validation and Open-Source: } We conduct experiments in single-session and multi-session to evaluate diverse and challenging scenarios \cite{kim2020mulran, barnes2020oxford, burnett2023boreas}.
                                                        Additionally, we aim to contribute to the advancement of the \ac{Radar} robotics community by sharing our code.
            
\end{itemize}

\section{Related Works}
Place recognition studies based on various sensors have shown remarkable results. 
Among them, camera and LiDAR sensors demonstrate high levels of performance.
Radar-based place recognition can be widely divided into System-on-Chip (SoC) \ac{Radar} and Imaging \ac{Radar}.
Additionally, it's worth noting that \ac{Sonar} images are similar to \ac{Radar} images.
Thus, in this section, we summarize place recognition research using each type of sensor.

\subsection{Place Recognition for other sensors} 
\subsubsection{Place Recognition for Camera}
Vision-based place recognition has been conducted for the longest time.
DBOW2 \cite{galvez2012bags} is a loop detection method that converts images into a bag of word vectors. 
In DBOW2, the perceptual aliasing problem is addressed and includes real-time geometric verification.
NetVLAD \cite{arandjelovic2016netvlad} represents the entire image as a compact single vector through CNN. 
NetVLAD has secured the flexibility to swap out features and developed an end-to-end architecture suitable for place recognition.

\subsubsection{Place Recognition for \ac{LiDAR}}
\ac{LiDAR}-based place recognition research is actively progressing in the robotics community. 
\textit{Scan context} \cite{kim2018scan} that utilizes structural information such as building height has shown powerful performance in urban environments. 
The \textit{Scan context} of 2D images contains semi-metric information and achieves rotational invariance.
The subsequent research algorithm, \textit{Scan context++} \cite{kim2021scan}, further enhances performance by achieving translational invariance in addition to rotational invariance. 
Recently, deep-learning-based research is gaining prominence. 
\textit{Logg3d-NET} \cite{vidanapathirana2022logg3dnet} leverages local features from \ac{LiDAR} to represent places more prominently.
Also, \textit{Overlaptransformer} \cite{ma2022overlap} utilizes range images in the training process to generate rotational invariant descriptors.

\subsubsection{Place Recognition for \ac{Sonar}}
Underwater research is at an earlier stage compared to \ac{Radar} sensors, but some studies have shown notable results. 
\citet{lee2014experimental} utilizes Forward Looking Sonar (FLS) to detect artificial markers, thereby enhancing localization performance. 
However, a drawback of this research is its limitation to place recognition only in man-made environments. 
Recently, \citet{kim2023robust} proposes a \textit{sonar context} that leverages structural information in the underwater environment.
This study focuses on understanding the meaningful information from the max intensity of \ac{Sonar} images and generating global descriptors similar to max pooling. 

\subsection{Place Recognition for Radar} 
\subsubsection{SoC Radar} SoC Radar is not suitable for place recognition because it uses a sparse pointcloud.
However, recent studies leveraging deep learning have shown noteworthy results in addressing this issue.
\textit{AutoPlace} \cite{cai2022autoplace} utilizes the \ac{Radar}'s velocity to remove dynamic objects and generate features through a spatial-temporal encoder.
\textit{mmPlace} \cite{meng2024mmplace} combines a rotating platform with the \ac{Radar} to create a heatmap and generates descriptors through a spatial encoder.

\subsubsection{Imaging Radar}
\citet{kim2020mulran} proposed the resized \textit{Scan context} to transform the radar image itself into a descriptor called \textit{Radar scan context} (RadarSC). 
This algorithm also achieves rotational invariance, aiding the \ac{ICP} algorithm with a semi-metric initial pose. 
\citet{jang2023raplace} utilizes the Radon Transform to generate rotational and translational invariant descriptors, \textit{Raplace}. 
Raplace has contributed to laying the foundation for \ac{Radar} place recognition.
\textit{Open-RadVLAD} (RadVLAD) \cite{gadd2024open} is a vector of locally aggregated descriptors that achieve rotational invariance. 
This algorithm has demonstrated its performance through an extensive amount of exhaustive results.
\section{Proposed Method}
\begin{figure}[h]
	\centering
	\def\width{0.48\textwidth}%
    {%
		\includegraphics[clip, trim= 10 90 10 60, width=\width]{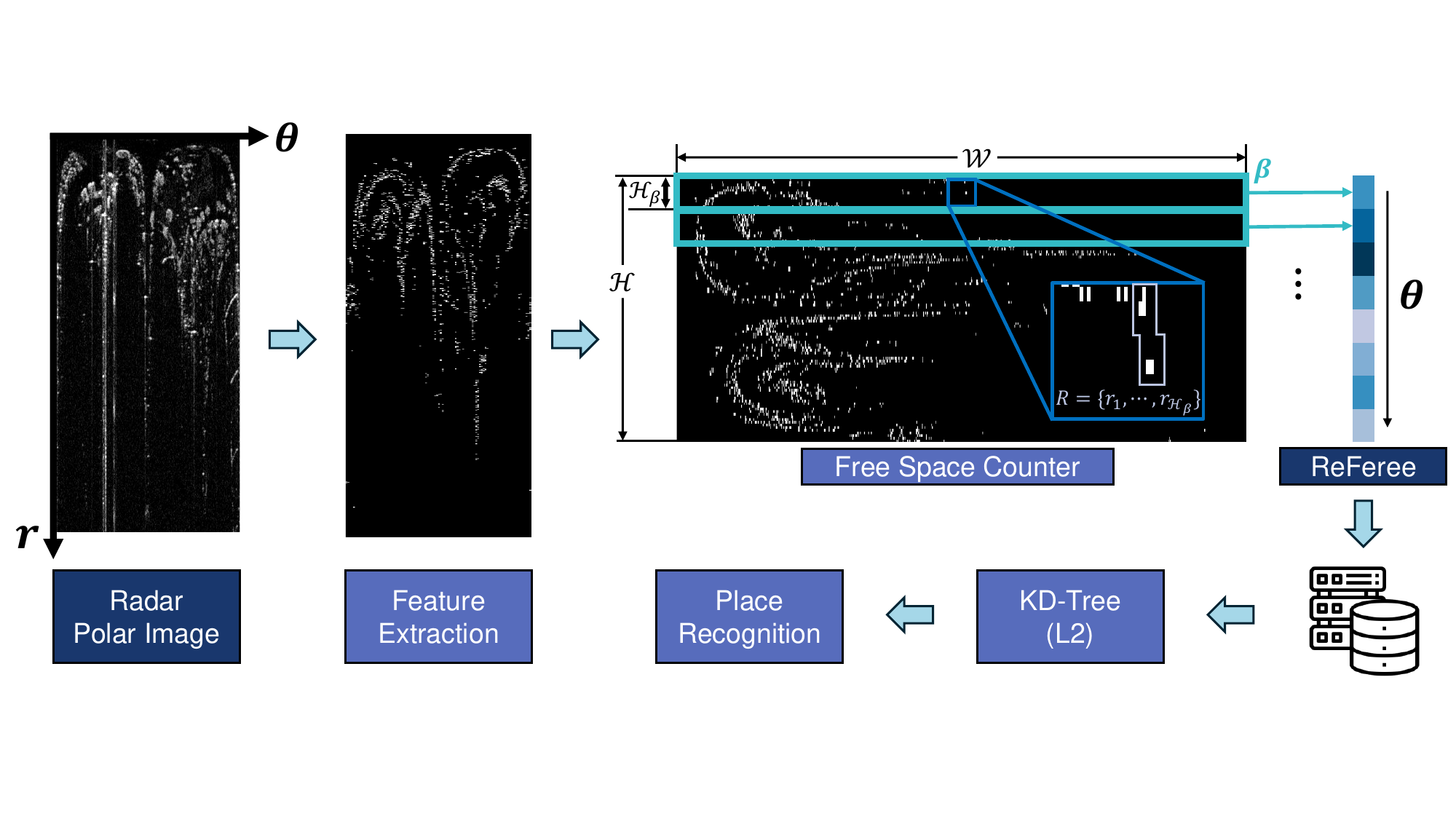}
	}
        \vspace{-0.5cm}
        \caption{Our proposed pipeline.
                 The navy rectangle represents data, and the blue rectangle represents the algorithm. 
                 After acquiring \ac{Radar} polar images from the sensor, we first proceed with feature extraction and counting free space to obtain the final descriptor called \textit{ReFeree}.
                 Second, we put the proposed descriptor in the database.
                 Finally, we build a KD-Tree to recognize revisited places through fast searching.
                 }
                 
        \vspace{-0.5cm}
	
 \label{fig:pipeline}
\end{figure}

As described in \figref{fig:pipeline}, our descriptor follows a simple pipeline.
The conversion of polar images from radar sensors into cartesian coordinate images is unnecessary.
Hence, techniques such as backward warping are not essential, thereby streamlining the description process.

\subsection{Feature Extraction} 
\ac{RCS} indicates a measure of an object's detectability by \ac{Radar}, and it varies depending on material properties and structural appearance.
\ac{Radar} image's intensity is heavily related to \ac{RCS}, it is proportional to the latent power multiplied by the \ac{RCS} of the target \cite{harlow2023new}.
Radar image define $\mathcal{R} \in\mathbb{R}^{\mathcal{H} \times \mathcal{W}}$ in polar coordinates, where $\mathcal{H}$ and $\mathcal{W}$ are the image's height and width, respectively.
The dimensions $\mathcal{H}$ and $\mathcal{W}$ are associated with angle and range, and encapsulating semi-metric information.

The primary component of our descriptor is how to handle feature information induced by intensity from the \ac{Radar} image.
We select the feature extraction algorithm proposed in \cite{cen2018precise} to distinguish valid signals from $\mathcal{R}$.
This algorithm involves decomposing the signal into high and low-frequency signals, 
 and integrating them according to a Gaussian scaling factor.
Finally, the integrated signal go through the thresholding process to extract valid features from the raw radar signal.
As parameters for feature extraction algorithm, we use threshold noise $z_q=6.0$ and variance of Gaussian smoothing $\sigma=17.0$ which operates in pixel metric space.
By using this method, we reduce noise from the signal and get higher clarity in classifying areas whether detected objects or vacancy.
We define the feature extracted Radar image as $\mathcal{R}_\mathcal{F} \in\mathbb{R}^{\mathcal{H} \times \mathcal{W}}$.


\subsection{Free space} 
Although feature extraction reduces noise and leaves only the relatively robust intensity, the features of \ac{Radar} sensors still have poor resolution, both geometrically and in terms of intensity, compared to denser sensors such as \ac{LiDAR}.
Inspired by \cite{maffei2020global}, this paper proposes a descriptor based on the vacant area (i.e. free-space) in $\mathcal{R}_\mathcal{F}$.
Free space is efficient due to its conciseness. 
Thus, the descriptor matching based on the overlap between free spaces is robust to the noisy and poor resolution characteristics of the \ac{Radar}.
By improving the methodology of \cite{maffei2020global} and embedding it as a high-dimensional vector, we reorganize it into a suitable representation for \ac{Radar}-based place recognition.


\begin{figure*}[t]
	\centering
	\def\width{0.98\textwidth}%
    {%
        \includegraphics[clip, trim= 0 60 80 0, width=\width]{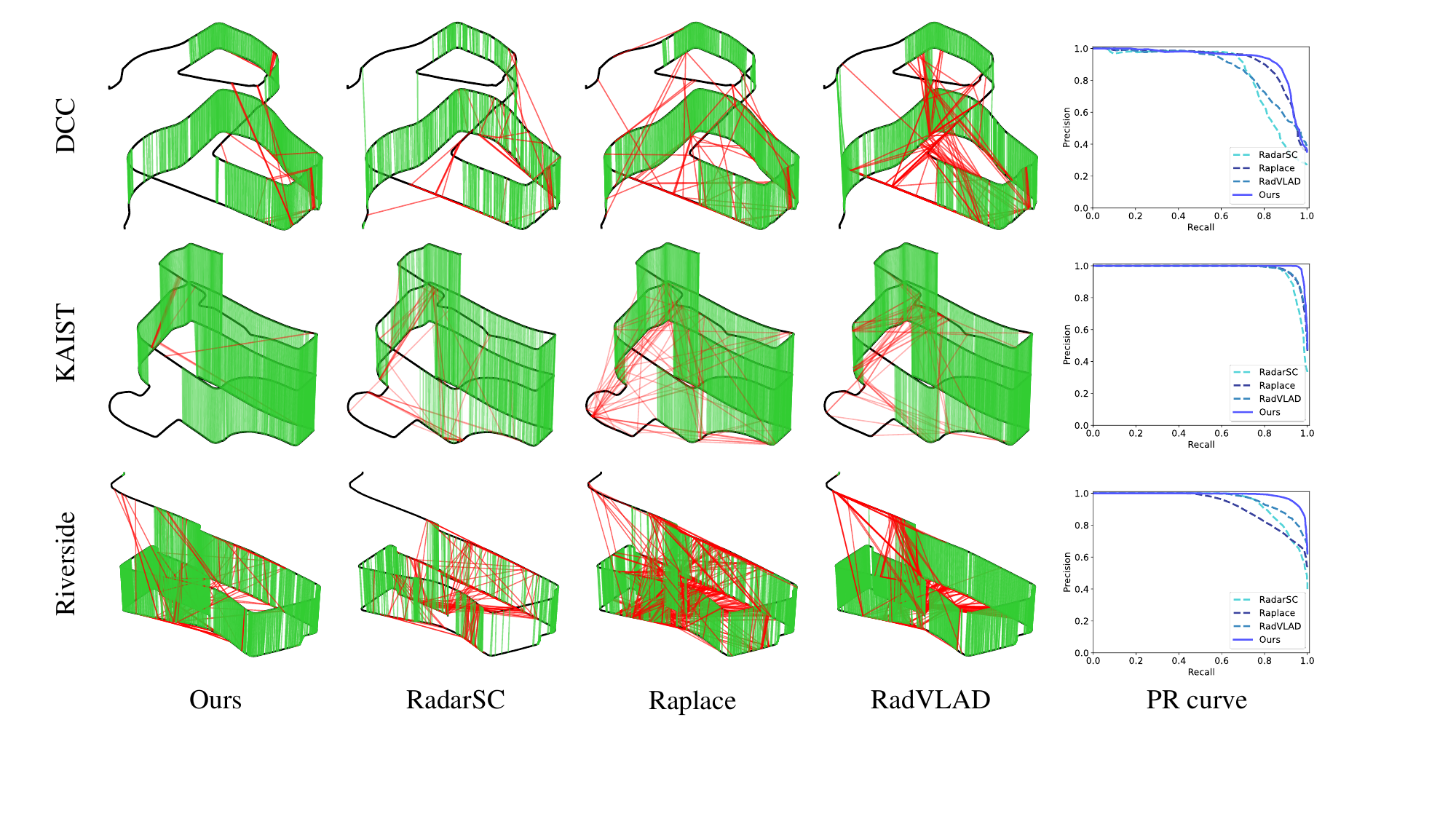}
    }
        \vspace{-0.2cm}
        \caption{Matching graph where F1 score is highest and a Precision-Recall (PR) curve in single session datasets.
                 In the matching graph, the green color represents true loops located within 20m.}
        \vspace{-0.5cm}
	
 \label{fig:single}
\end{figure*}

\subsection{ReFeree} 
Using the features and the free space, we propose a \ac{Radar}-based efficient global descriptor called \textit{ReFeree}.
To transform $\mathcal{R}_\mathcal{F}$ into a \textit{ReFeree} $\mathbf{K}$, which is an $\alpha$-dimensional vector that each of the elements represents a compression of range-wise block $\beta$'s free space information, we represents $\mathbf{K}$ by:
\begin{equation}
	\label{equ:free space counter}
    \mathbf{K} = \{K_{1}, \cdots, K_{\alpha}\}
\end{equation}
where $\beta \in \mathbb{R}^{\mathcal{H}_\beta \times \mathcal{W}}$ is subsection that compose of $\mathcal{R}_\mathcal{F}$, and $\mathcal{H}_\beta$ is $\mathcal{H}$ divided by $\alpha$.
Free space density of each range-wise block $K_{\beta}$ is computed as:
\begin{equation}
	\label{equ:free space block}
    K_{i} = \frac{\sum_{j=1}^{\mathcal{\mathcal{H}_\beta}}
    \left( \sum_{k=1}^{r_j} s(b_{jk}) \right)}{\mathcal{H}_\beta \cdot \mathcal{W}}
\end{equation}
where $r_j$ is element of farthest feature index vector $R = \{r_1, \cdots, r_{\mathcal{H}_\beta}\}$, and it is matched with each row of $\beta$. The notation $b_{jk}$ is a state of single pixel located at $j^{th}$ row, $k^{th}$ column in $\beta$, can be converted to numeric value by state function $s(b)$:
\begin{equation}
	\label{equ:free or not}
    s(b) = 
    \begin{cases}
        1, & \mbox{if }b\mbox{ is a \textit{free-space}} \\
        0, & \mbox{otherwise}
    \end{cases}
\end{equation}

\subsection{Place Retrieval} 
For retrieving place from candidates $\mathcal{C}$, we need to establish similarity distance and searching algorithm.
Given that \textit{ReFeree} $\mathbf{K}$ is an $\alpha$-dimensional vector, the KD-Tree algorithm is well-suited for our descriptor, considering the efficiency and speed in searching for distances and optimal candidate $c^{*}$.
From this algorithm, we can compute similarity distance $d_s$ in the descriptor space, and translational distance $d_t$ using the position of the $c^{*}$.
Both distances can be calculated using the L2-norm formula and have similarity threshold $\tau_s$ and translational threshold $\tau_t$, respectively. Whole retrieval algorithm for query descriptor $\mathbf{K}^{q}$ can be formulated as:
\begin{equation}
	\label{equ:free space counter}
    c^{*} = \underset{c \in \mathcal{C}}{\argmin}\;\mathbf{D}(\mathbf{K}^{q}, \mathbf{K}^{c}),\;\;s.t.\;\;\mathbf{D}(\mathbf{K}^{q},  \mathbf{K}^{c}) < \mathbf{T}
\end{equation}
where $\mathbf{D}$ is function that calculate a set of distance between $\mathbf{K}^{q}$ and $\mathbf{K}^{c}$, which can represent $\mathbf{D}(\mathbf{K}^{q},  \mathbf{K}^{c})=\{d_s,d_t\}$, and $\mathbf{T}$ is a set of distance threshold, $\mathbf{T}=\{\tau_s,\tau_t\}$.
Only pairs that satisfy both thresholds respectively are treated as detected loops.
\section{Experimental Results}

We validate our descriptor \textit{ReFeree} in various datasets.
The datasets used for validating single-session are part of Mulran \cite{kim2020mulran} that includes a variety of sequences. 
DCC offers structural diversity and KAIST enables testing of various dynamic objects and reverse loops, while Riverside features two bridges that create structural repetitions in certain sections.  
For multi-session validation, we evaluate with Oxford Offroad Radar Dataset (OORD) \cite{gadd2024oord}, Oxford Radar RobotCar \cite{barnes2020oxford}, and Boreas \cite{burnett2023boreas}.
OORD captures a naturistic and rugged environment that contains unpaved terrain, and sloped trails with diverse weather conditions such as snowfall across several periods.
Boreas has suitable sequences for multi-session evaluation but lacks loops.
On the other hand, Oxford Radar RobotCar contains sufficient loops for place recognition, and it is also appropriate for multi-session evaluations.
All experiments are conducted using Python code on an Intel i7-12700 KF.

\begin{figure*}[t]
	\centering
	\def\width{0.98\textwidth}%
    {%
        \includegraphics[clip, trim= 0 20 50 0, width=\width]{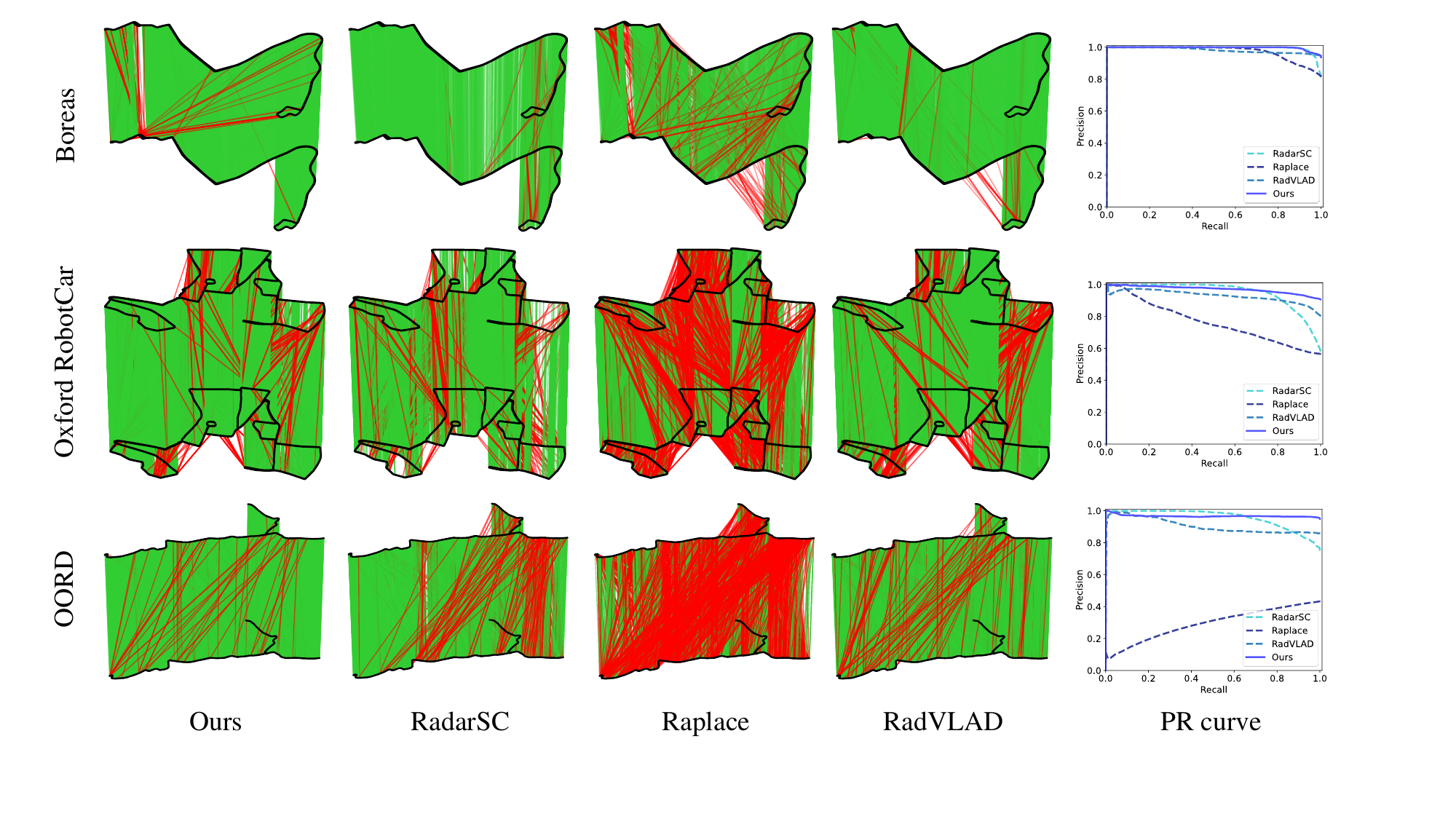}
	}
        \vspace{-0.8cm}
        \caption{Matching graph where F1 score is highest and a Precision-Recall (PR) curve in multi-session datasets.
                 In the matching graph, the green color represents true loops located within 20m.}
	
 \label{fig:multi}
\end{figure*}

\begin{table*}[t]
\caption{Evaluation Results}
\renewcommand{\arraystretch}{1.2}
\centering\resizebox{\textwidth}{!}{
{\tiny
\begin{tabular}{cc|cccccc|cccccc}
\toprule
\hline
\multicolumn{2}{c|}{Session}                          & \multicolumn{6}{c|}{Single session}                                                                    & \multicolumn{6}{c}{Multi session}                                                                  \\ \hline
\multicolumn{2}{c|}{Datasets}                         & \multicolumn{2}{c|}{DCC}          & \multicolumn{2}{c|}{KAIST}        & \multicolumn{2}{c|}{Riverside} & \multicolumn{2}{c|}{OORD}         & \multicolumn{2}{c|}{Boreas}       & \multicolumn{2}{c}{Oxford RobotCar} \\ \hline
\multicolumn{2}{c|}{Metric}                           & AUC                 & \multicolumn{1}{c|}{F1 max}               & AUC                 & \multicolumn{1}{c|}{F1 max}  
                                                      & AUC                 & F1 max                                    & AUC                 & \multicolumn{1}{c|}{F1 max} 
                                                      & AUC                 & \multicolumn{1}{c|}{F1 max}               & AUC                 & F1 max        \\ \hline

\multicolumn{1}{c|}{\multirow{7}{*}{Method}} & RadarSC    & 0.908               & \multicolumn{1}{c|}{0.793}                & 0.975               & \multicolumn{1}{c|}{0.927} 
                                                      & \textbf{0.950}      & 0.852                                     & \bl{\textbf{0.892}} & \multicolumn{1}{c|}{0.790}  
                                                      & \bl{\textbf{0.960}} & \multicolumn{1}{c|}{0.916}                & \bl{\textbf{0.914}} & 0.799              \\

\multicolumn{1}{c|}{}                        & Raplace     & \textbf{0.927}      & \multicolumn{1}{c|}{\textbf{0.849}}       & 0.982               & \multicolumn{1}{c|}{0.936} 
                                                      & 0.866               & 0.815                                     & 0.226               & \multicolumn{1}{c|}{0.604}  
                                                      & 0.880               & \multicolumn{1}{c|}{0.905}                & 0.691               & 0.722              \\

\multicolumn{1}{c|}{}                        & RadVLAD     & 0.884               & \multicolumn{1}{c|}{0.778}                & \textbf{0.983}      & \multicolumn{1}{c|}{\textbf{0.937}} 
                                                      & 0.937               & \textbf{0.885}                            & 0.587               & \multicolumn{1}{c|}{\textbf{0.924}}  
                                                      & 0.747               & \multicolumn{1}{c|}{\bl{\textbf{0.973}}}  & \textbf{0.793}      & \textbf{0.901}              \\

\multicolumn{1}{c|}{}                        & RadarSC-DS & 0.546               & \multicolumn{1}{c|}{0.113}                & 0.765               & \multicolumn{1}{c|}{0.366} 
                                                      & 0.784               & 0.508                                     & 0.604               & \multicolumn{1}{c|}{0.133}  
                                                      & 0.692               & \multicolumn{1}{c|}{0.253}                & 0.565               & 0.104              \\

\multicolumn{1}{c|}{}                        & Raplace-DS  & 0.773               & \multicolumn{1}{c|}{0.553}                & 0.902               & \multicolumn{1}{c|}{0.766} 
                                                      & 0.866               & 0.730                                     & 0.201               & \multicolumn{1}{c|}{0.321}  
                                                      & 0.680               & \multicolumn{1}{c|}{0.585}                & 0.565               & 0.324              \\

\multicolumn{1}{c|}{}                        & RadVLAD-DS  & 0.714               & \multicolumn{1}{c|}{0.450}                & 0.720               & \multicolumn{1}{c|}{0.481} 
                                                      & 0.686               & 0.588                                     & 0.549               & \multicolumn{1}{c|}{0.694}  
                                                      & 0.589               & \multicolumn{1}{c|}{0.442}                & 0.564               & 0.357              \\

\multicolumn{1}{c|}{}                        & Ours   & \bl{\textbf{0.941}} & \multicolumn{1}{c|}{\bl{\textbf{0.881}}}  & \bl{\textbf{0.993}} & \multicolumn{1}{c|}{\bl{\textbf{0.976}}} 
                                                      & \bl{\textbf{0.979}} & \bl{\textbf{0.941}}                       & \textbf{0.645}      & \multicolumn{1}{c|}{\bl{\textbf{0.976}}}  
                                                      & \textbf{0.946}      & \multicolumn{1}{c|}{\textbf{0.971}}       & 0.785               & \bl{\textbf{0.951}}             \\ \hline \bottomrule
\end{tabular}}}
\vspace{-0.5cm}
\end{table*}

\subsection{Comparison Descriptor}
As represented in \tabref{table:desc}, we compare our method with RadarSC, Raplace, and RadVLAD.
Compared to other descriptors used in our experiments, the proposed method is at least 3$\times$ and up to 497$\times$ lighter and performs efficiently relative to the information represented.
Hence, we validate the efficiency of the proposed method by assessing its performance when other methods possess an equivalent amount of information (shape) as ours in Section \uppercase\expandafter{\romannumeral4}-C.
The RadarSC and RadVLAD approaches resize the incoming \ac{Radar} images to fit our descriptor, while Raplace adjusts the downsampling rate to achieve a similar capacity to the proposed method.
The downsampled methods are denoted as RadarSC-DS, Raplace-DS, and RadVLAD-DS, respectively.

\begin{table}[h]
\caption{Descriptor Size Comparision}
\renewcommand{\arraystretch}{1.2}
\centering\resizebox{0.48\textwidth}{!}{
\begin{tabular}{c|c|c|c|c|c|c|c}
\hline
\hline
Method              & RadarSC          & Raplace            & RadVLAD             & RadarSC-DS        & Raplace-DS       & RadVLAD-DS         & Ours         \\ \hline
Descriptor size [B] & 1616         & 3008          & 262272         & 346           & 576         & 5248          & 528          \\ \hline
Descriptor shape    & 20$\times$60  & 20$\times$18  & 32768$\times$1 & 5$\times$10   & 8$\times$7  & 640$\times$1  & 50$\times$1  \\ \hline \hline
\end{tabular}}
\label{table:desc}
\end{table}
\vspace{-0.2cm}

\subsection{Evaluation Metrics}
\subsubsection{PR curve}
For evaluating the performance of place recognition, the Precision-Recall (PR) curve is utilized. 
Precision and recall are defined as below:
\begin{equation}
\text { Precision }=\frac{\mathrm{TP}}{\mathrm{TP}+\mathrm{FP}}, \quad \text { Recall }=\frac{\mathrm{TP}}{\mathrm{TP}+\mathrm{FN}},
\end{equation}
where TP, FP, and FN are true positive, false positive, and false negative, respectively. 

\subsubsection{F1 score}
F1 score is the harmonic mean of precision and recall, which means the performance of the classifier.
The F1 score metric is defined as follows:
\begin{equation}
\text { F1 score}=\frac{2 \times \mathrm{Precision} \times \mathrm{Recall}}{\mathrm{Precision}+\mathrm{Recall}},
\end{equation}
where precision and recall are mentioned above, respectively. 

\subsubsection{AUC score}
We also evaluate the AUC score, which represents the area under the ROC curve consisting of True Positive Rate (TPR) and False Positive Rate (FPR).
TPR and FPR are defined as below:
\begin{equation}
\text { FPR }=\frac{\mathrm{FP}}{\mathrm{FP}+\mathrm{TN}}, \quad \text { TPR }=\frac{\mathrm{TP}}{\mathrm{TP}+\mathrm{FN}},
\end{equation}
where TP, FP, and TN, FN are true positive, false positive, and true negative, false negative, respectively. 

\subsection{Evaluation of Place Recognition}
To validate the place recognition performance of our method, we evaluate two cases of localization scenarios (i.e. single session and multi-session).
The results of place recognition are \figref{fig:single}, \figref{fig:multi}, and the matching graphs represent true matching and false matching of the highest F1 score sequence.
Since RadarSC is not officially published and open-sourced, we evaluated it by implementing the same validation code as the Mulran dataset.

\subsubsection{Single Session Validation}
In the validation of a single session, the most important characteristic of the dataset is abundant loops, and the following datasets are suitable to the mentioned conditions - DCC, KAIST, and Riverside of the Mulran dataset.
As depicted in \figref{fig:single}, our method detects a significantly higher number of true loops and fewer false loops compared to the other methods.
PR curves represented in \figref{fig:single} show superior classification performance in whole tested datasets.

\subsubsection{Multi Session Validation}
The validation of multi-session is meaningful to checking robustness to spatial-temporal changes and can be utilized in applications like map-merging or multi-robot mapping.
\figref{fig:multi} shows the matching graph and PR curve of 3 datasets - OORD for the unstructured environment, Oxford RobotCar, and Boreas for the structured environment.
For validation purposes, we use sequences with a 4 to 8-day interval for structured environment datasets and a 30-day interval for unstructured environments where environmental changes are relatively minimal.
Similar to the results of the single session, the matching graph and PR curve are both dominant compared to the other methods.
In the PR curve of OORD, our method maintains high performance while Raplace shows low performance due to sensor model differences with other datasets.
\section{Conclusion}
We propose a lightweight and efficient \ac{Radar}-based global descriptor called \textit{Referee}.
We validate the superiority of our method across datasets from various scenarios.
However, our method still has few problems to solve like reverse loop or validation by our own dataset.
In future work, we plan to enhance our descriptor for \ac{SLAM} by incorporating rotation invariance.

%


\bibliographystyle{packages/IEEEtranN} 
\bibliography{packages/string-short, packages/references}

\end{document}